\documentclass[journal,twoside]{IEEEtran}

\usepackage{graphicx}
\usepackage{cite}
\usepackage{multirow}
\usepackage{amsmath}
\usepackage{array}
\usepackage{tabu}
\usepackage{flushend}
\ifCLASSOPTIONcompsoc
  \usepackage[caption=false,font=normalsize,labelfont=sf,textfont=sf]{subfig}
\else
  \usepackage[caption=false,font=footnotesize]{subfig}
\fi
\makeatletter
\def\hlinewd#1{%
  \noalign{\ifnum0=`}\fi\hrule \@height #1 \futurelet
   \reserved@a\@xhline}
\makeatother
\begin{document}

\title{Rectifier Neural Network with a Dual-Pathway Architecture for Image
Denoising}

\author{Keting Zhang
        and Liqing Zhang
\thanks{The work was supported by the National Natural Science Foundation of China (Grant No. BC0300355), the National Basic Research Program of China (Grant No. 2015CB856004) and the Key Basic Research Program of Shanghai (Grant No. 15JC1400103).}
\thanks{The authors are with Key Laboratory of Shanghai Education Commission for Intelligent Interaction and Cognitive Engineering, Department of Computer Science and Engineering, Shanghai Jiao Tong University, Shanghai, 200240 (e-mail: zzsnail@sjtu.edu.cn; zhang-lq@cs.sjtu.edu.cn)}
}
\markboth{}
{Zhang and Zhang: Rectifier Neural Network with a Dual-Pathway Architecture for Image
Denoising}

\maketitle

\begin{abstract}
Recently deep neural networks based on tanh activation function have shown their impressive power in image denoising.
In this letter, we try to use rectifier function instead of tanh and propose a dual-pathway rectifier neural network by combining two rectifier neurons with reversed input and output weights in the same hidden layer. We drive the equivalent activation function and compare it to some typical activation functions for image denoising under the same network architecture. The experimental results show that our model achieves superior performances faster especially when the noise is small.
\end{abstract}


\IEEEpeerreviewmaketitle

\section{Introduction}
\IEEEPARstart{T}{he} target of image denoising is to recover the original clean image under the additive white Gaussian noise corruption. Many classical patch-based algorithms which exploit natural image statistics exist, such as sparse redundant representation model \cite{elad2006image,sahoo2015enhancing,feng2015optimized} and non-local statistics model \cite{dabov2007image,mairal2009non,sun2013analysis}. These methods are often well-engineered and some are widely considered as the current state-of-the-art, e.g. BM3D algorithm \cite{dabov2007image}.

Different from the above methods, recently the machine learning approach based on deep neural network has draw considerable attention. This approach utilizes neural network to approximate a denoising function from a noisy patch to a clean patch. Some network models have been proposed for this task, including stacked sparse autoencoder \cite{xie2012image}, convolutional neural network \cite{jain2009natural} and plain neural network \cite{burger2012image}. It has been shown that a multi-layer enormous plain neural network trained on a large training set is able to achieve comparable denoising performance to BM3D method \cite{burger2012image}. Usually, larger network and larger training set can lead to better results \cite{burgerthesis2013}. In addition, some investigations have also been made to improve specific texture denoising \cite{wang2016note}, or to provide robustness to different noise types \cite{agostinelli2013adaptive} or noise levels \cite{wang2014can}.

Although different architectures are adopted, the models mentioned above are all based on traditional tanh activation function. Whereas, recent emerging work has shown that non-saturating rectifier function performs better \cite{nair2010rectified,glorot2011deep} and trains faster \cite{krizhevsky2012imagenet} than saturating tanh on image recognition task. In this letter, we study the use of rectifier function in deep neural network for image denoising. We choose the plain feed-forward neural network architecture because it has been shown to achieve state-of-the-art performance.

As pointed out in \cite{glorot2011deep}, rectifier is a one-sided function which means its response to the opposite of a strongly excitatory input is zero. By empirically analyzing the orthogonality of the dictionary learned by rectifier neurons, we find that they tend to learn the reversed atoms due to the one-sided property. To remove such redundancy, we propose a dual-pathway architecture by combining two rectifier neurons with reversed input and output weights in the same hidden layer.
This strategy results in an equivalent antisymmetric activation function which enables one node to respond to patterns with opposite polarities simultaneously. Thus there is greater chance to update the weights from two paths, which benefits the training of large neural networks on large datasets. We have already successfully used this model for non-blind image deconvolution \cite{zhang2017non}.

Then we evaluate the proposed activation function against some typical non-linear units (e.g., sigmoid, tanh, rectifier and parametric rectifier \cite{he2015delving}) in deep network framework for image denoising. We make the comparisons on different noise levels to observe the performance progress during the model training. The experimental results show that our method outperforms all competitors. The improvements are significant especially when the noise is relatively small. Compared to other activation functions, our model achieves superior performances much faster. We provide a Matlab toolbox with the trained models to test our approach.

\section{Dual-Pathway Rectifier Neural Network}
\subsection{Motivation}
\begin{figure*}[ht]
  \centering
  \subfloat[]{
    \includegraphics[scale=0.365]{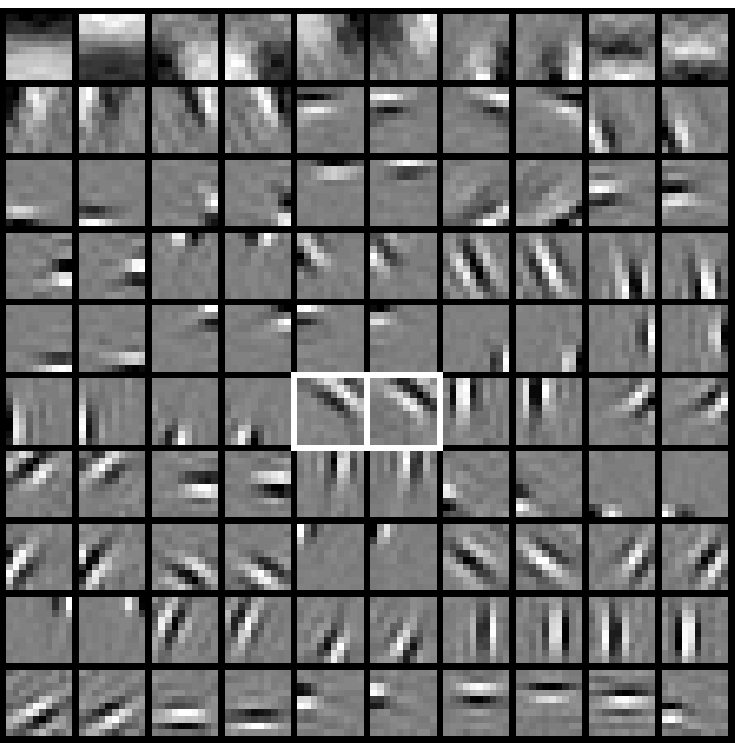}}
  \subfloat[]{
    \includegraphics[scale=0.327]{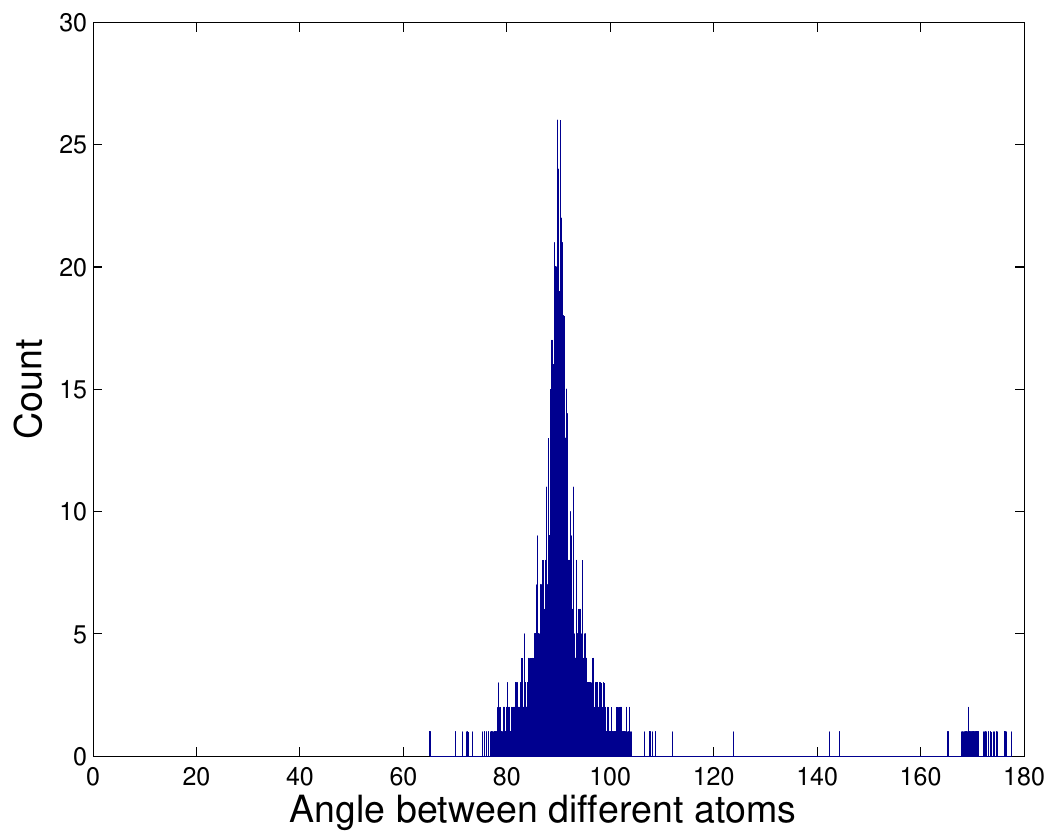}}
  \subfloat[]{
    \includegraphics[scale=0.327]{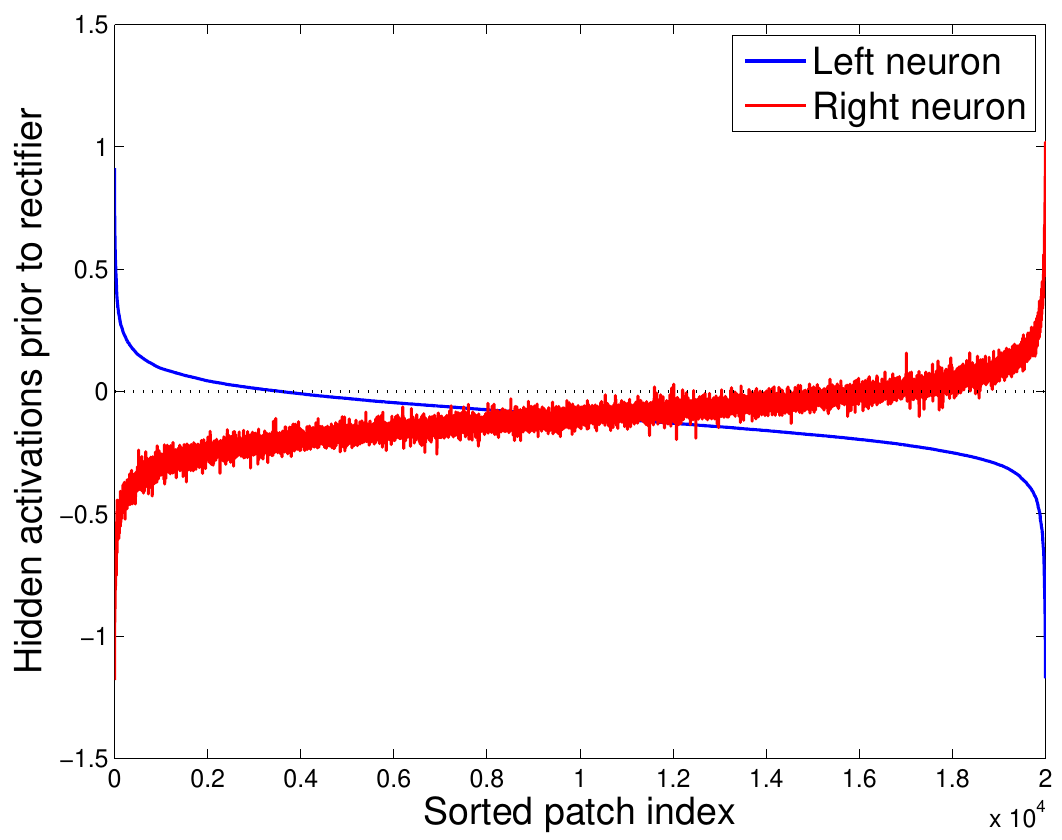}}
  \caption{The effect of one-sided property on the dictionary learned by rectifier neural network. (a) all learned atoms are shown in pairs and sorted according to the angles between atoms in each pair in descending order. (b) histogram of angles between different atoms. (c) hidden activations prior to rectifier nonlinearity of two neurons corresponding to the atoms in white boxes in (a). All test patches are sorted by the activations of neuron with the left atom.}
\end{figure*}
First we examine qualitatively the effect of rectifier's one-sided property on the orthogonality of the dictionary learned by a single hidden layer network. The network takes a corrupted version $\tilde{x}$ of original clean patch $x$ as input and maps it to a denoised patch $y$ by:
\begin{equation}
  y=r(\tilde{x})=W_{2}f(W_{1}\tilde{x}+b_{1})+b_{2}
\end{equation}
where $W_{1}$ and $W_{2}$ are weight matrices, $b_{1}$ and $b_{2}$ are bias vectors, and $f=max(0,x)$ is the rectifier activation function. To illustrate the learned atoms better, we constrain $W_{1}$ and $W_{2}$ to be transposes of each other.

We extract some image patches of size $10\times10$ and corrupt them using additive white Gaussian noise with $\sigma=25$ to form the corresponding noisy patches. To eliminate the effect of DC component, all patches are preprocessed by subtracting the mean-value for each sample. We use a network with 100 hidden units to learn a dictionary consisting of 100 atoms.

In Fig. 1(a), the learned atoms are shown in pairs and sorted according to the angles between them in descending order.
We can see that most pairs show opposite patterns at the same spatial location. The histogram of all angles between different atoms is plotted in Fig. 1(b), which shows that most angles are around 90 degrees, while a few are relatively large. These relatively large angles exactly correspond to atom pairs in (a). It is indicated that the atoms in different pairs are quite orthogonal to each other, while atoms in the same pair show reversed polarities.

We feed some test patches into the model to observe the responses of hidden neurons. The hidden activations prior to rectifier nonlinearity of two neurons corresponding to the atoms in white boxes in (a) are shown in Fig. 1(c). The activations of right neuron fluctuate because these two atoms are not exactly opposite to each other. This figure clearly shows that many opposite patterns exist in natural images, while the rectifier neuron which can be activated by some pattern does not respond to the corresponding opposite pattern because of its one-sided property. Thus for rectifier neurons, some redundancies exist in the learned dictionary. Next we will propose a dual-pathway architecture to improve it.

\subsection{Model Description}
The basic idea of our model is that for every rectifier neuron in hidden layers, we add an extra companion node and associate it with the opposite input and output weights. Specifically, for every rectifier node connected with input weight $w_{in}$ and output weight $w_{out}$, we generate a companion node in the same layer with input weight $-w_{in}$ and output weight $-w_{out}$. By this strategy, later we will show that these two neurons are equivalent to one neuron with a novel activation function. The purpose of associating with $w_{in}$ and $-w_{in}$ is to enable the equivalent neuron to respond to both opposite patterns in data, while $w_{out}$ and $-w_{out}$ is to reflect the polarity of input patterns correctly.

A single hidden layer network with such architecture is shown in Fig. 2. The mapping defined by it is:

\begin{equation}
\begin{split}
y&=\begin{bmatrix}W_{2} & -W_{2}\end{bmatrix}f\left(\begin{bmatrix}W_{1} \\ -W_{1}\end{bmatrix}\tilde{x}+\begin{bmatrix}b_{1} \\ b_{1}'\end{bmatrix}\right)+b_{2}\\
\end{split}
\end{equation}
where $\tilde{x}$ is the noisy patch, $y$ is the obtained denoised patch, the weight matrices $W_{1}$, $W_{2}$ and biases $b_{1}$, $b_{1}'$, $b_{2}$ are the parameters, $f$ is rectifier function $max(0,x)$.
\begin{figure}[t]
\centering
\includegraphics[scale=0.5]{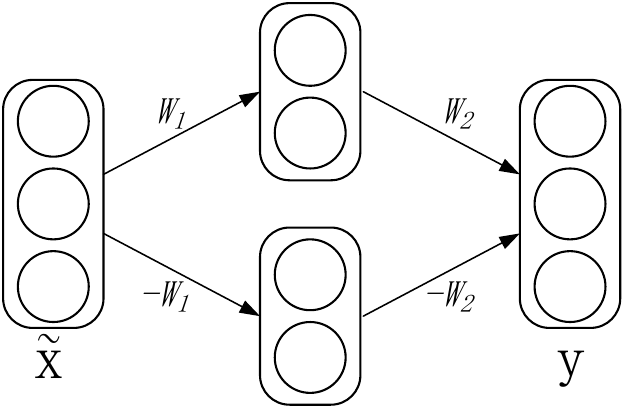}
\caption{Architecture of dual-pathway rectifier neural network with single hidden layer. For every neuron in hidden layer, there is an extra companion node with opposite input and output weights. The lower hidden nodes can be seen as the companions of the upper ones.}
\end{figure}

\begin{figure*}[t]
  \centering
  \subfloat[]{
    \includegraphics[scale=0.365]{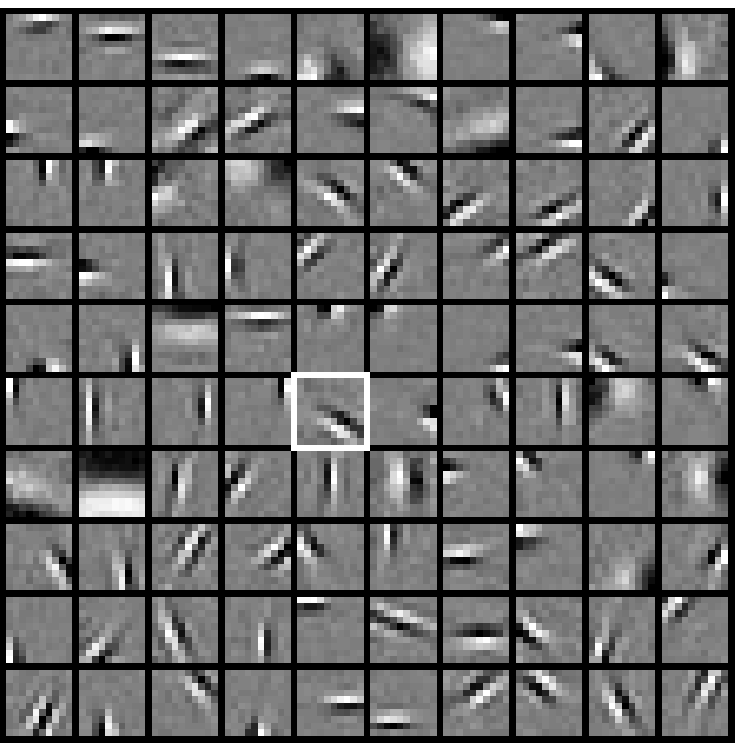}}
  \subfloat[]{
    \includegraphics[scale=0.327]{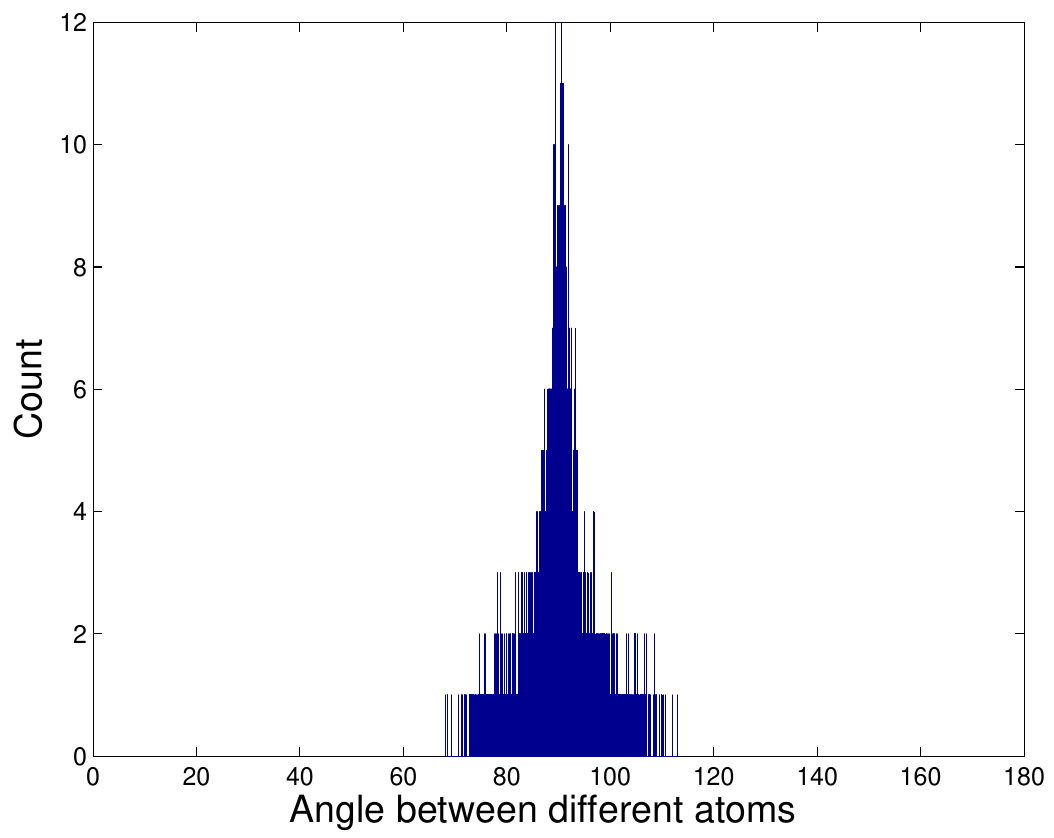}}
  \subfloat[]{
    \includegraphics[scale=0.327]{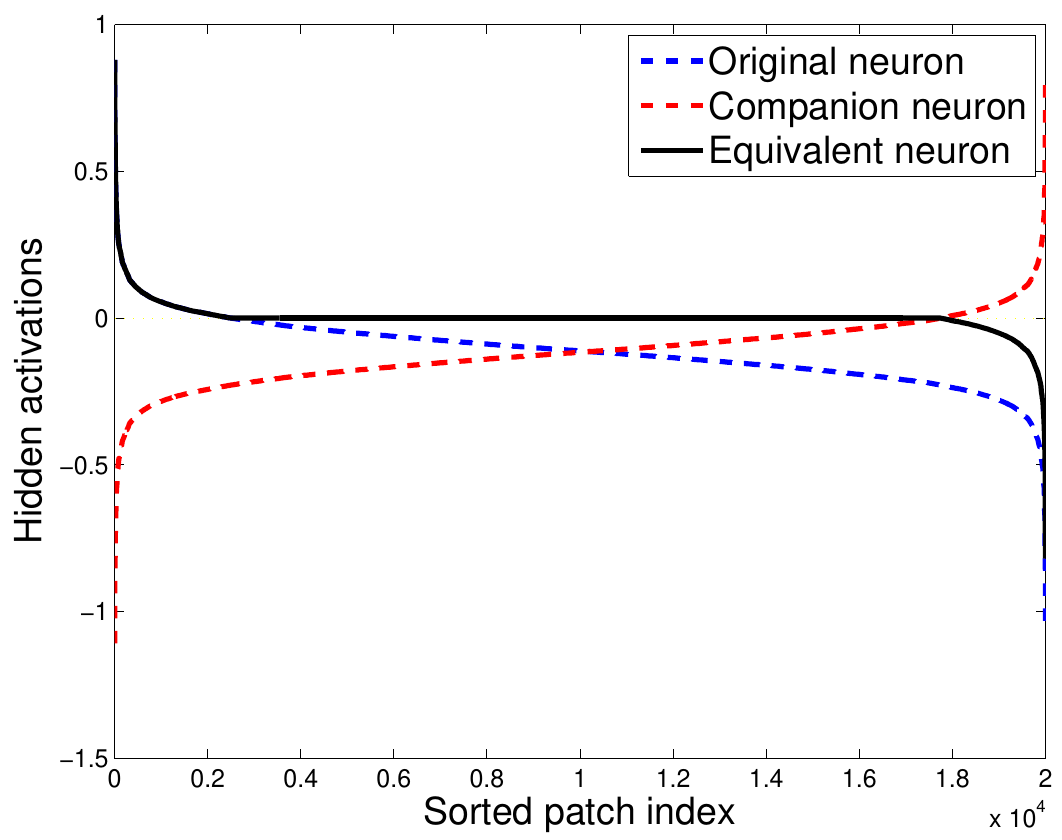}}
  \caption{Dictionary learned by a dual-pathway rectifier network. (a) all atoms are shown in the same way as Fig. 1(a). (b) histogram of angles between different atoms. (c) hidden activations prior to rectifier of original neuron (corresponding to the atom in white box in (a)) and the corresponding companion neuron, and hidden activations after rectifier of the equivalent neuron. All test patches are sorted by the activations of original neuron in descending order.}
\end{figure*}
We define a novel activation function $g(x)$ with parameter $t$ as follows:
\begin{equation}
g(x)=max(0,x+t)-max(0,-x+t)
\end{equation}
Hence, if we set $t=(b_{1}+b_{1}')/2$, then
\begin{equation}
\begin{split}
y&=W_{2}\left[f\left(W_{1}\tilde{x}+b_{1}\right)-f\left(-W_{1}\tilde{x}+b_{1}'\right)\right]+b_{2}\\
&=W_{2}g(W_{1}\tilde{x}+\frac{b_{1}-b_{1}'}{2})+b_{2}
\end{split}
\end{equation}
Thus we can see that the dual-pathway architecture is equivalent to the antisymmetric activation function $g(x)$ with trainable parameter $t$. Its shapes with non-negative and negative parameters are shown in Fig. 3.
\begin{figure}[t] 
\centering
\subfloat{
\includegraphics[scale=0.26]{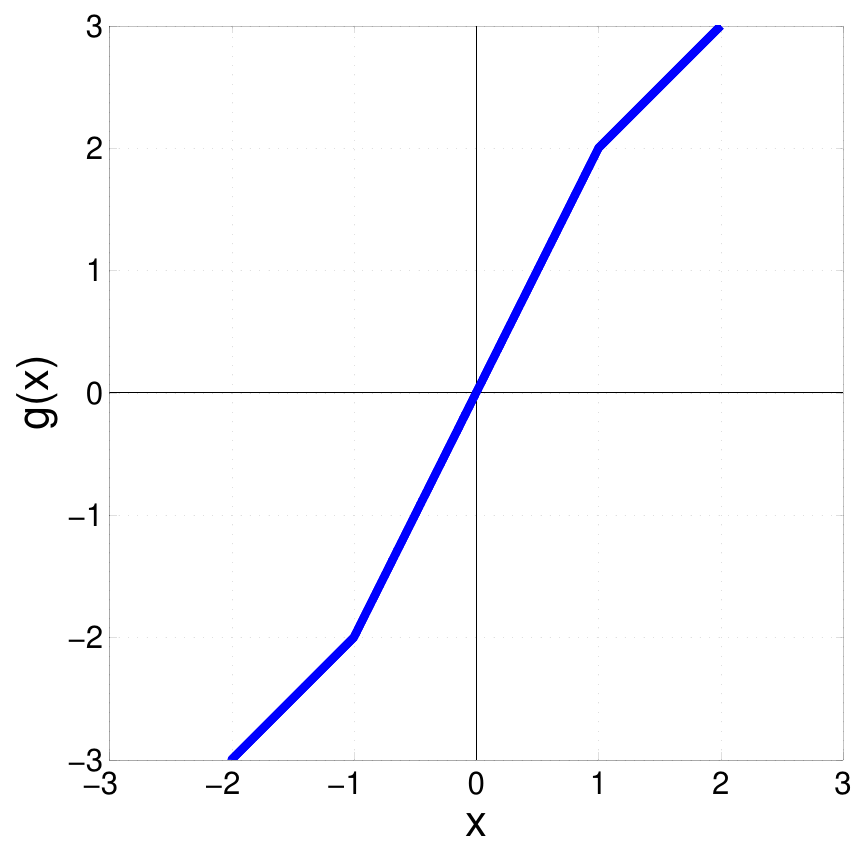}}
\subfloat{
\includegraphics[scale=0.26]{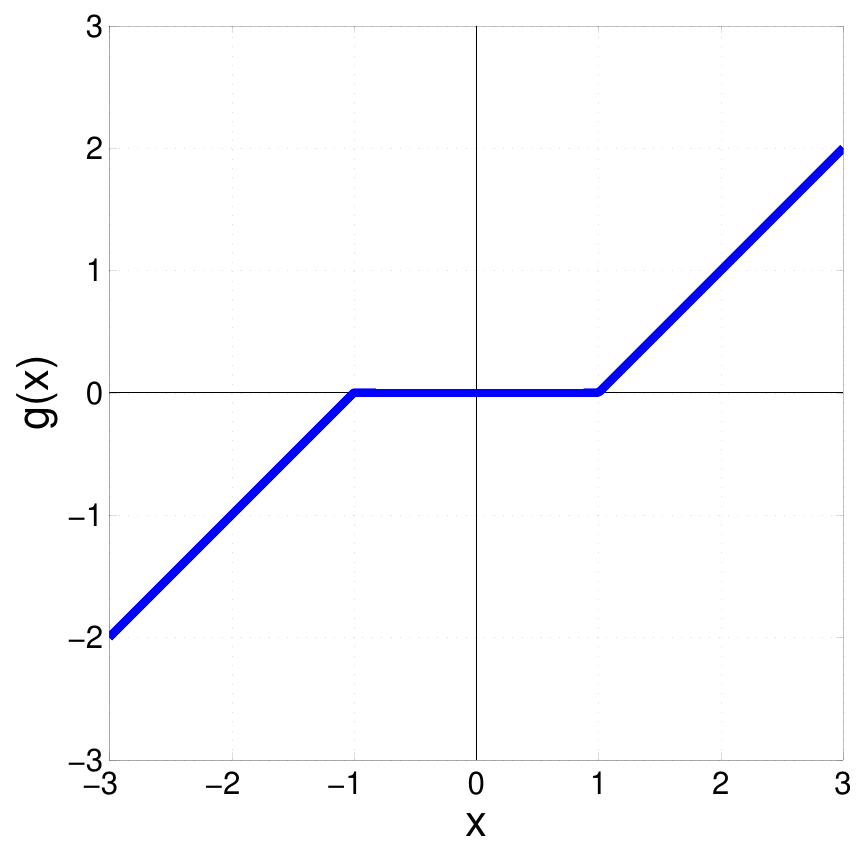}}
\caption{Activation function $g(x)$ with trainable parameter $t$. \textbf{Left:} $t$ is non-negative ($t=1$). \textbf{Right:} $t$ is negative ($t=-1$).}
\end{figure}

\subsection{Application to image denoising}
First we train a deep neural network on many pairs of clean and the corresponding noisy patches to learn a denoising function. Following \cite{burger2012image}, the network takes large noisy patch as input and tries to recover its central small block as output to obtain slight performance gain. All parameters including weights, biases and $t$s are
learned simultaneously using back-propagation algorithm \cite{lecun1989backpropagation}, minimizing the empirical squared error between obtained denoised patches and the clean patches.
The derivative of $g(x)$ with respect to $t$ is given by:
\begin{equation}
  \frac{\partial g(x)}{\partial t}=
  \begin{cases}
    0 & x\leq|t|\\
    sgn(x)  & x>|t|
  \end{cases}
\end{equation}
In this paper, all network parameters are optimized using minibatch Limited memory BFGS (L-BFGS) method which is very suitable for training deep models \cite{ngiam2011optimization}. We use L-BFGS implementation in minFunc by Mark Schmidt\footnote{http://www.cs.ubc.ca/~schmidtm/Software/minFunc.html}.

During the testing phase, given a noisy image, we first chop it into a number of overlapping noisy patches. Then we apply our trained network to them and get the denoised patches. Finally, all denoised patches are put at the positions of their noisy counterparts and aggregated on the overlapping regions via Gaussian weighted averaging.

\section{Experimental Study}

\subsection{Dictionary Learning}
We carry out the similar experiment to the one in Section
\uppercase\expandafter{\romannumeral2}-A to illustrate the effect of proposed dual-pathway model on dictionary learning. As shown in Fig. 4(a), we try to display all learned atoms in the same way as Fig. 1(a). However, we can see that none of atom pairs exhibit opposite patterns, which is very different from the plain rectifier model.
Fig. 4(b) presents the histogram of all angles between different atoms, which demonstrates that all atoms are quite orthogonal to each other. 

Then we randomly choose a atom in the white box in Fig. 4(a) and depict its hidden activations in Fig. 4(c). We plot the activations prior to rectifier nonlinearity of original neuron and its companion neuron using dashed lines. We can see that they detect the opposite patterns in input respectively, which is similar to plain rectifier network. Whereas the equivalent neuron combines their functions and yields responses according to the polarities of the input. This mechanism enables one single neuron to encode both opposite patterns simultaneously.

The results of this comparative experiment suggest two advantages of dual-pathway architecture. One is the removal of the redundancy in the learned dictionary caused by rectifier's one-sided property. Whereas, experimentally we find that such redundancy becomes less serious when the dictionary is highly overcomplete. This might be due to the fact that greater flexibility is provided in capturing structure from the data in higher dimension. Thus for one specific pattern, its representation is on longer unique and there is no need for a neuron which exactly matches it. The other advantage is that because both opposite patterns can activate one neuron, there is greater chance to update the corresponding weights. This may promote the efficiency of model training.

\subsection{Natural Image Denoising}
\begin{figure*}[t]
\begin{center}
    \includegraphics[scale=0.24]{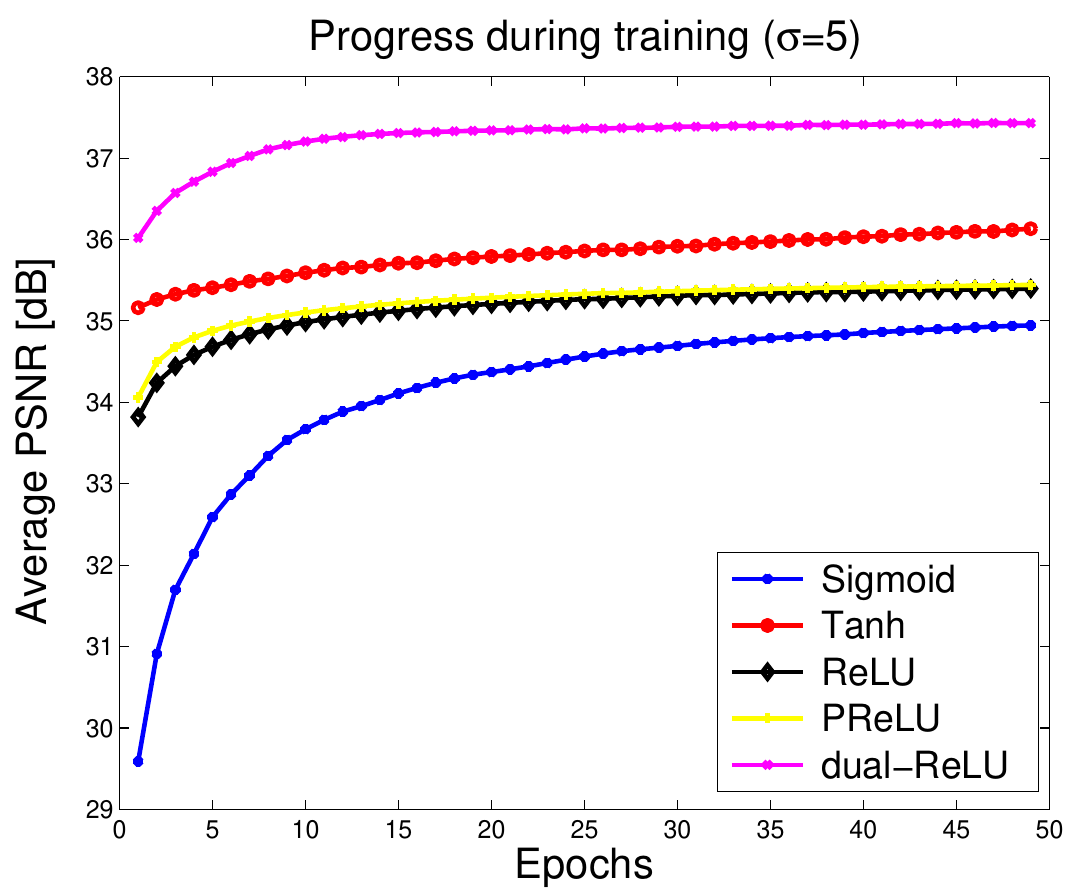}
    \includegraphics[scale=0.24]{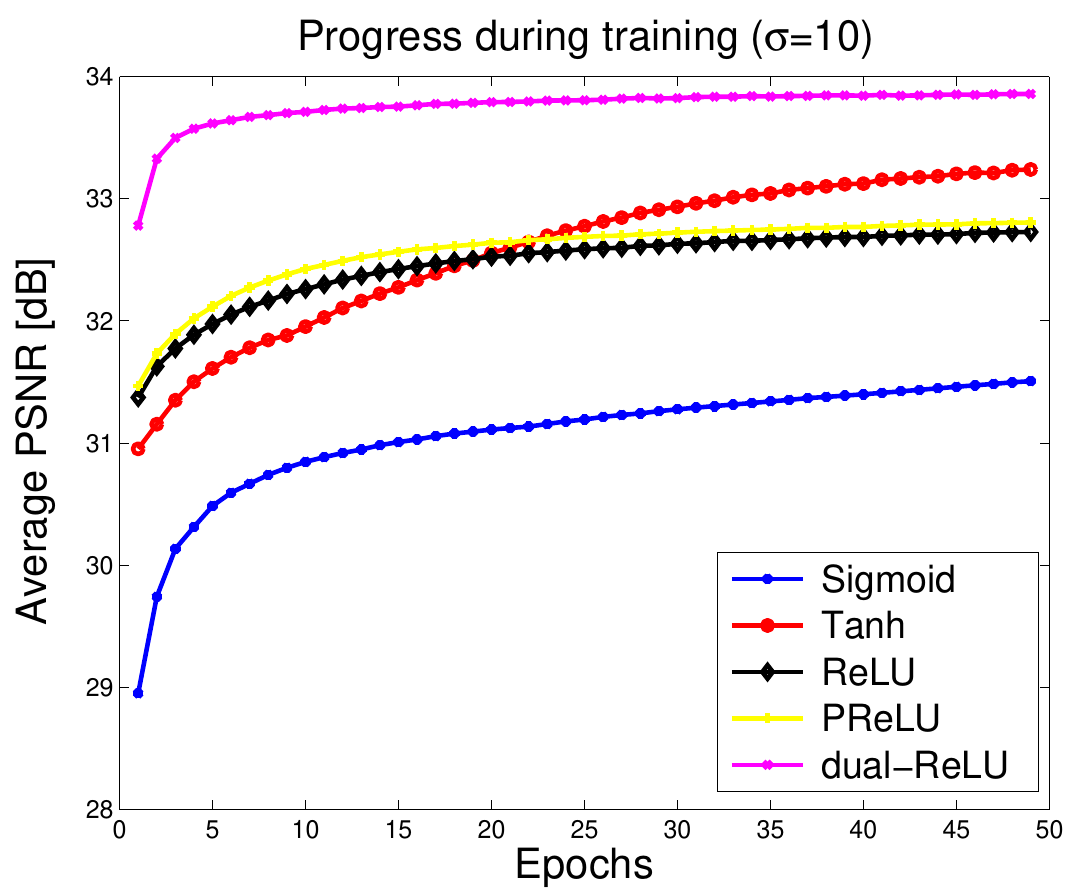}
    \includegraphics[scale=0.24]{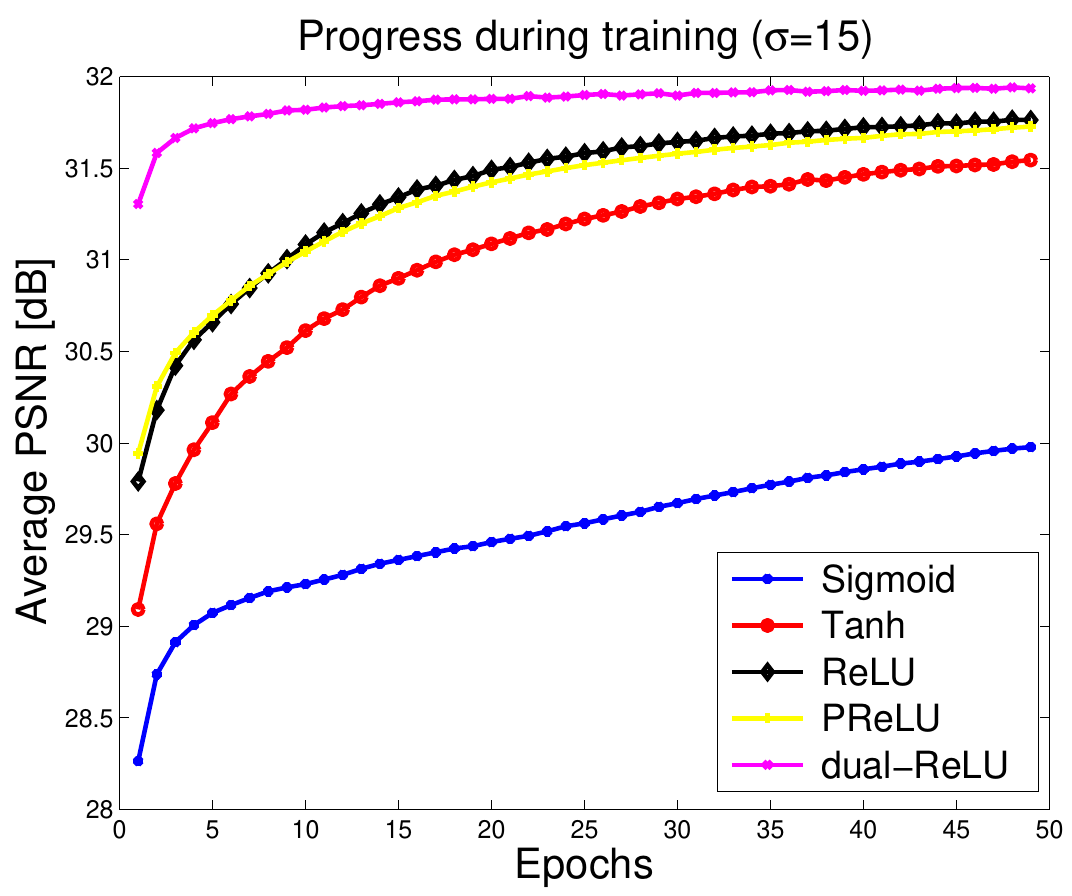}
    \includegraphics[scale=0.24]{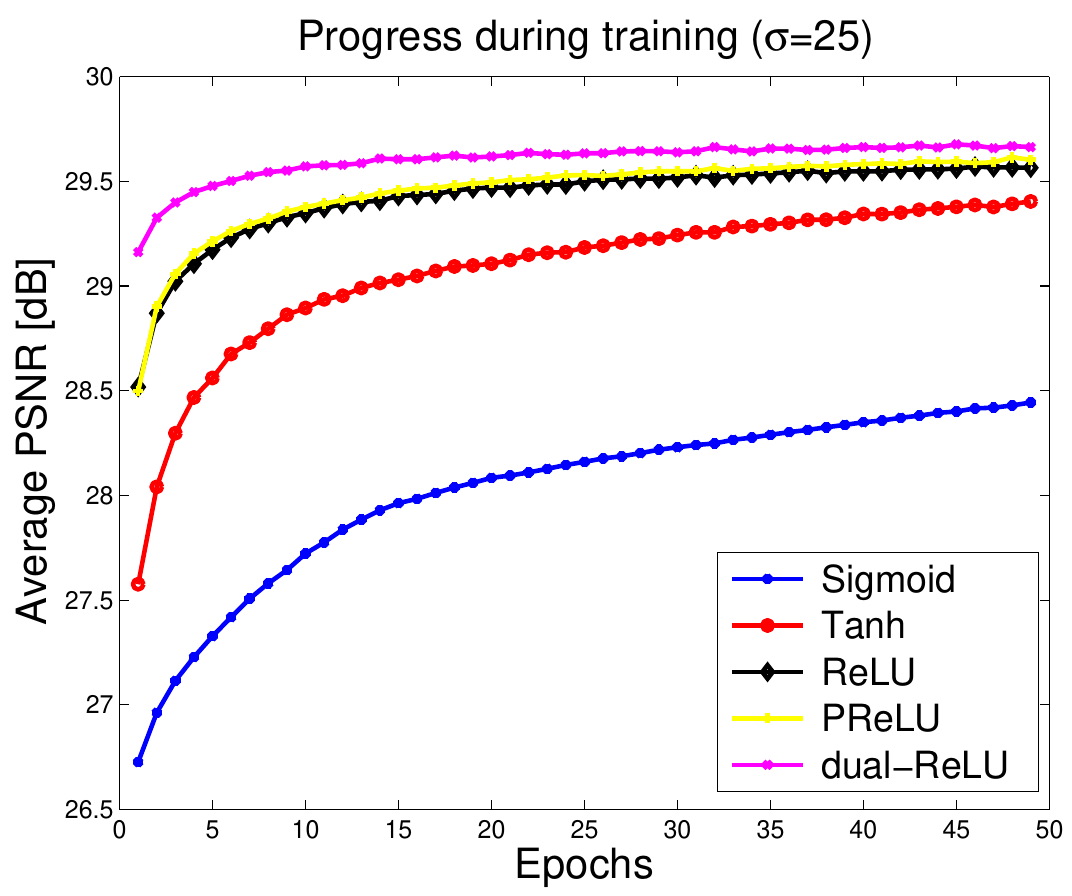}
\end{center}
\begin{center}
    \includegraphics[scale=0.24]{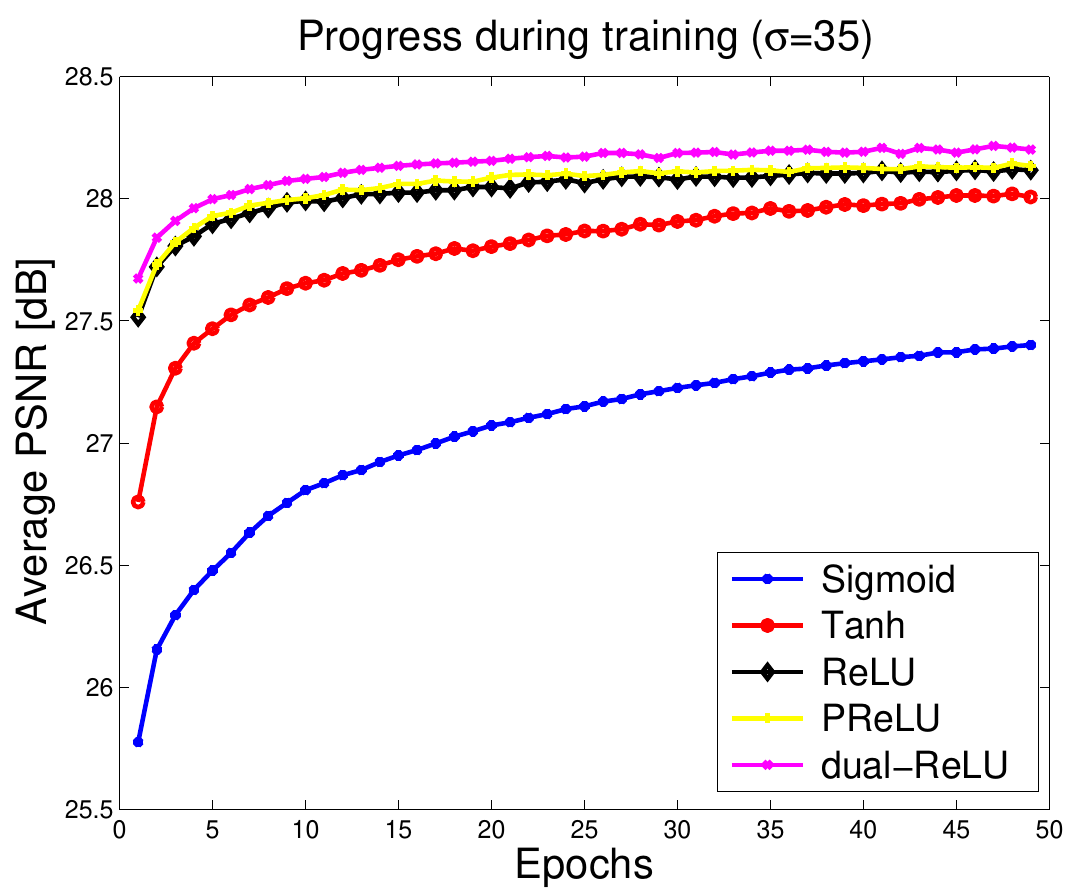}
    \includegraphics[scale=0.24]{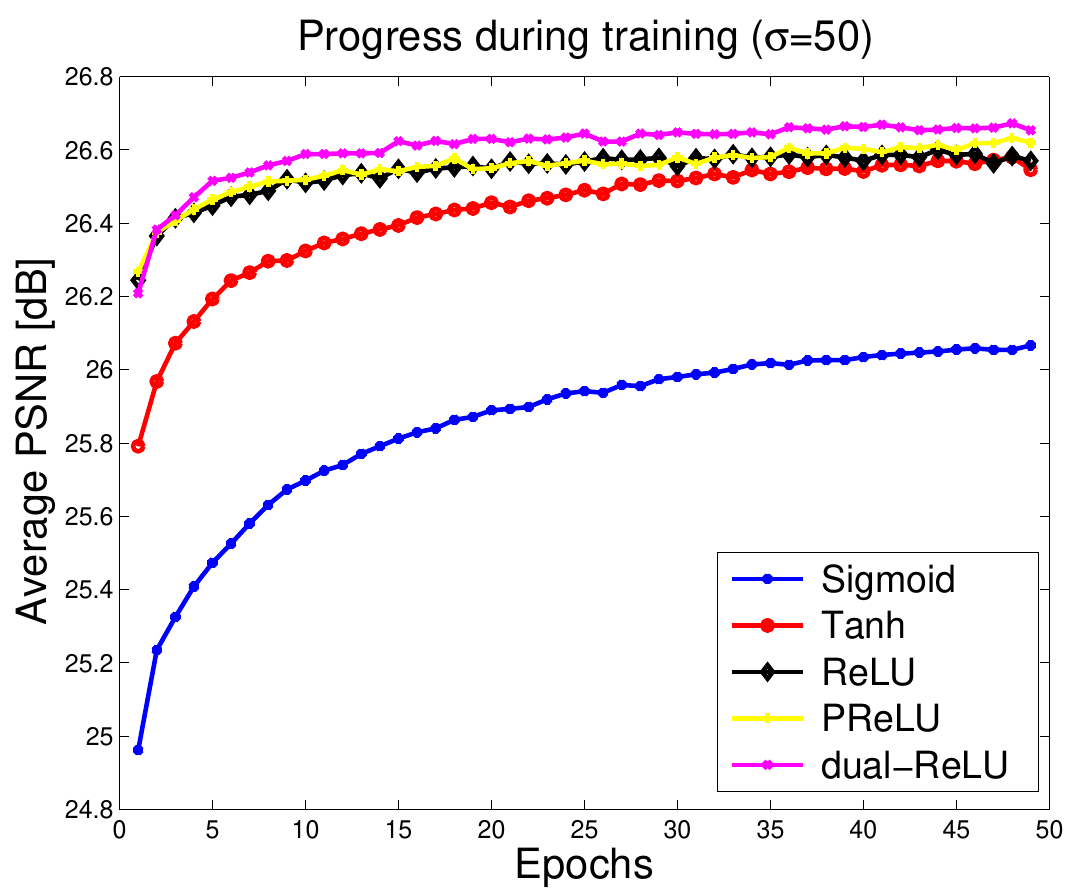}
    \includegraphics[scale=0.24]{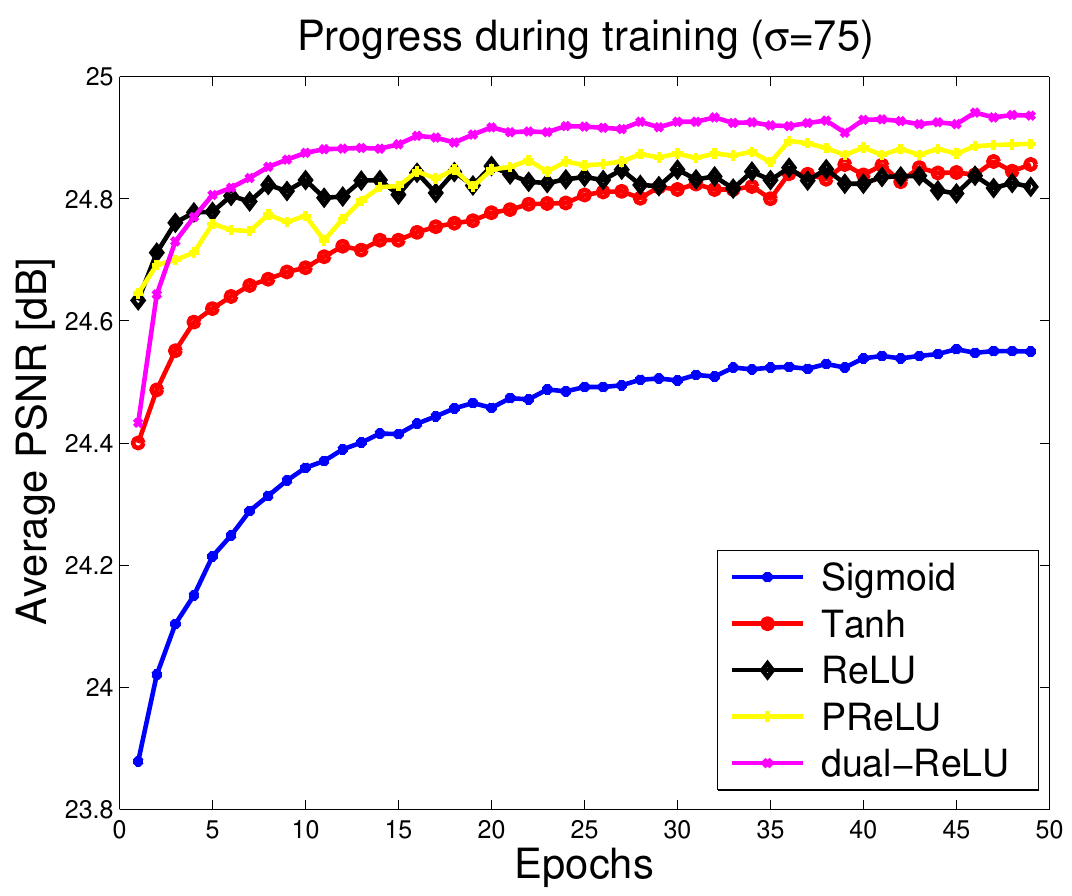}
    \includegraphics[scale=0.24]{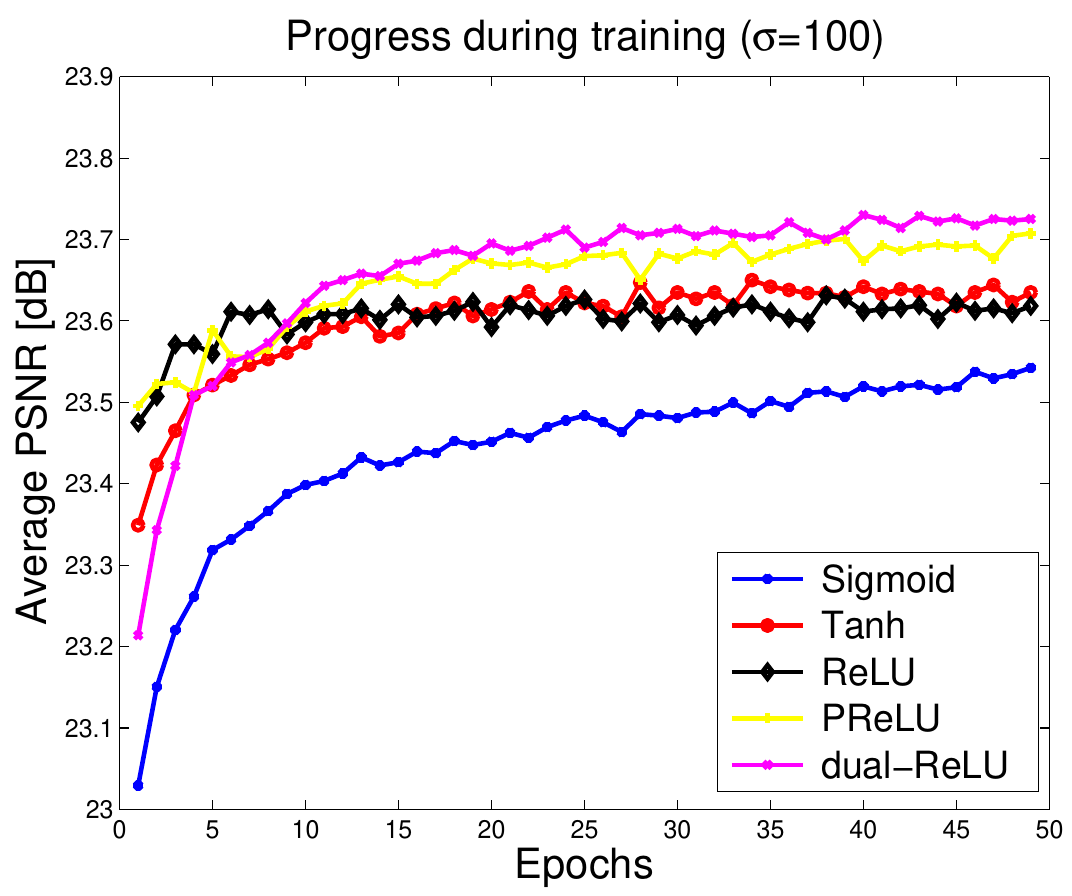}
\end{center}
  \caption{Improving average PSNR on the test images for eight Gaussian noise levels during model training.}
\end{figure*}

\begin{table}[h]
\centering   
\caption{Comparisons of the denoising performances by PSNR.}
\label{my-label}
\renewcommand{\arraystretch}{1.3}  
\begin{tabular}{c|ccccc}
\hlinewd{1pt} \rule{0pt}{14pt}
Noise  & Sigmoid & Tanh  & ReLU  & PReLU & dual-ReLU      \\ [3pt]\hline
5      & 34.95   & 36.13 & 35.40 & 35.44 & \textbf{37.43} \\
10     & 31.51   & 33.24 & 32.73 & 32.81 & \textbf{33.86} \\
15     & 29.98   & 31.54 & 31.76 & 31.73 & \textbf{31.94} \\
25     & 28.44   & 29.40 & 29.56 & 29.60 & \textbf{29.66} \\
35     & 27.40   & 28.01 & 28.12 & 28.13 & \textbf{28.20}  \\
50     & 26.07   & 26.55 & 26.57 & 26.62 & \textbf{26.65} \\
75     & 24.55   & 24.86 & 24.82 & 24.89 & \textbf{24.94} \\
100    & 23.54   & 23.64 & 23.62 & 23.71 & \textbf{23.73} \\ \hlinewd{1pt}
\end{tabular}
\end{table}

We adopt a deep network architecture with four hidden layers of size 512, an input layer of size 289 and an output layer of size 81. This network takes a $17\times17$ noisy patch as input and tries to recover its central $9\times9$ block. The dictionary learned in the output layer is 6 times larger than the denoised patch.
We use the natural images in the Berkeley segmentation database \cite{martin2001database} and convert them to gray-scale images to generate the training patches. The test set consists of eight standard images: Boat, Bridge, Cameraman, Couple, Hill, Lena, Man and Peppers. Eight levels of additive white Gaussian noise with standard deviations $5$, $10$, $15$, $25$, $35$, $50$, $75$ and $100$ are tested. For every specific noise level, we generate 50 million training samples and choose the minibatch size 10000. We run L-BFGS algorithm with a minibatch for 50 iterations and resample a new minibatch. The training of the model requires about 2 days of computation time on single Tesla K20c GPU.

We compare the proposed dual-pathway rectifier (dual-ReLU) with some commonly used activation functions including sigmoid, tanh, rectifier (ReLU) and parametric rectifier (PReLU). Quantitative comparisons are performed using average Peak Signal to Noise Ratio (PSNR).
The denoising performances of all models are summarized in Table \uppercase\expandafter{\romannumeral1}. We can see that the dual-ReLU model achieves the best results for all noise levels.
\begin{figure}[t]
\begin{center}
\includegraphics[scale=0.29]{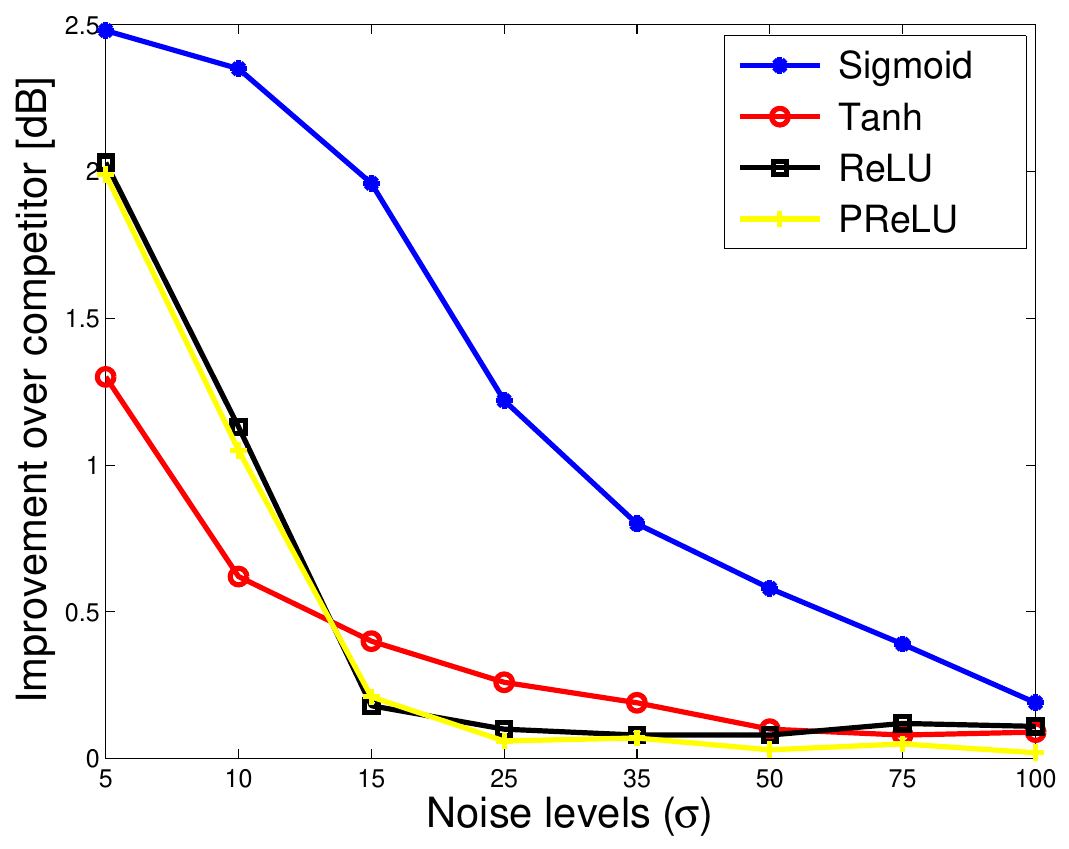}
\caption{The performance improvements in PSNR of our method over some typical activation functions for different noise levels.}
\end{center}
\end{figure}

We plot the performance improvements of dual-ReLU model over other functions, as shown in Fig. 6. When the noise is small, the improvements are significant. As the noise increases, the improvements become small. Compared to tanh, rectifier and parametric rectifier, the results of our method are slightly better at high noise levels. To monitor the performance progress, we test all models after every one million training examples. Fig. 5 shows the improving average PSNR for all noise levels during model training.

In all cases, our models reach the steady states with superior performances faster than the competitors. Especially when the noise is low, in terms of training time with gradient descent, the dual-pathway rectifier function is much faster than the saturating sigmoid or tanh nonlinearities and single pathway rectifier or parametric rectifier neurons. At high noise levels, the differences in training efficiency are small. The presumable reason is that the bigger noise in data provides stronger gradients, which facilitate the model training. The results confirm the effectiveness of dual-pathway architecture in image denoising, which is consistent with previous analysis.

\section{Conclusion}
In this paper, we have proposed a dual-pathway rectifier network for image denoising. It reduces the redundancy in learned dictionary and improves the training efficiency.
For future work, we would like to investigate the application of dual-pathway architecture in convolutional neural network, not only for image restoration but for recognition task as well.

\bibliographystyle{IEEEtran}
\bibliography{IEEEabrv,bare_revised}
\end{document}